\documentclass[10pt,twocolumn,letterpaper]{article}

\usepackage{cvpr}
\usepackage{times}
\usepackage{graphics}
\usepackage{graphicx}
\usepackage{amsmath}
\usepackage{amssymb}
\usepackage{longtable,booktabs,array}
\usepackage{color}
\usepackage{subfigure}
\usepackage{multirow}
\usepackage{algorithm}
\usepackage{algorithmic}
\usepackage{epsfig}
\usepackage{verbatim}
\usepackage{paralist}

\usepackage[breaklinks=true,bookmarks=false]{hyperref}

\cvprfinalcopy 

\ifcvprfinal\fi
\begin{document}

\title{Fixed-point Factorized Networks}
\author{
Peisong Wang$^{1,2}$ and Jian Cheng$^{1,2,3} \thanks{The corresponding author.}$\\
$^1$Institute of Automation, Chinese Academy of Sciences \\
$^2$University of Chinese Academy of Sciences\\
$^3$Center for Excellence in Brain Science and Intelligence Technology, CAS \\
{\tt\small \{peisong.wang, jcheng\}@nlpr.ia.ac.cn}
}

\maketitle
\thispagestyle{empty}

\begin{abstract}
In recent years, Deep Neural Networks (DNN) based methods have achieved 
remarkable performance in a wide range of tasks and have been among the 
most powerful and widely used techniques in computer vision.
However, DNN-based methods are both computational-intensive and 
resource-consuming, which hinders the application of these methods on 
embedded systems like smart phones. To alleviate this problem, we introduce 
a novel Fixed-point Factorized Networks (FFN) for pretrained models to reduce the
computational complexity as well as the storage requirement of networks. 
The resulting networks have only weights of -1, 0 and 1, which 
significantly eliminates the most
resource-consuming multiply-accumulate operations (MACs).
Extensive experiments on large-scale ImageNet classification task show the
proposed FFN only requires one-thousandth of multiply operations with comparable accuracy.

\end{abstract}

\section{Introduction}

Deep neural networks (DNNs) have recently been setting new state of the art 
performance in many fields including computer vision, speech recognition
as well as natural language processing. Convolutional neural networks (CNNs), 
in particular, have outperformed traditional machine learning algorithms on 
computer vision tasks such as image recognition, object detection, semantic 
segmentation as well as gesture and action recognition. These breakthroughs 
are partially due to the added computational complexity and the storage 
footprint, which makes these models very hard to train as well as to deploy. 
For example, the Alexnet \cite{krizhevsky2012imagenet} 
involves 61M floating point parameters and 725M 
high precision multiply-accumulate operations (MACs). Current DNNs are 
usually trained offline by utilizing specialized hardware like NVIDIA GPUs and
CPU clusters. But such an amount of computation may be unaffordable for 
portable devices such as mobile phones, tablets and wearable devices, which 
usually have limited computing resources. What's more, the huge storage 
requirement and large memory accesses may hinder efficient hardware 
implementation of neural networks, like FPGAs and neural network 
oriented chips.

To speed-up test-phase computation of deep models, lots of matrix and tensor 
factorization based methods are investigated by
the community recently \cite{denil2013predicting, jaderberg2014speeding, zhang2015accelerating, 
lebedev2014speeding, kim2015compression, wang2016accelerating}. 
However, these methods commonly utilize full-precision weights, 
which are hardware-unfriendly especially for embedded systems. Moreover, 
the low compression ratios hinder the applications of these methods on 
mobile devices.

Fixed-point quantization can partially alleviate these two problems mentioned above.
There have been many studies working on reducing the storage and the 
computational complexity of DNNs by quantizing the parameters of these models. 
Some of these works \cite{courbariaux2014low, dettmers20158, gupta2015deep, lin2015fixed, miyashita2016convolutional}
quantize the pretrained weights using several bits 
(usually 3$\sim$12 bits) with a minimal loss of performance. 
However, in these kinds of quantized networks one still needs to employ large numbers
of multiply-accumulate operations. Others  
\cite{lin2015neural, cheng2015training, courbariaux2015binaryconnect, 
courbariaux2016binarynet, kim2016bitwise, hubara2016binarized, rastegari2016xnor}
focus on training these networks 
from scratch with binary (+1 and -1) or ternary (+1, 0 and -1) weights.
These methods do not rely on pretrained models and may reduce the computations
at training stage as well as testing stage. But on the other hand, these
methods could not make use of the pretrained models very efficiently 
due to the dramatic information loss during the binary or ternary quantization 
of weights. 

In this paper, we propose a unified framework called Fixed-point Factorized 
Network (FFN) to simultaneously 
accelerate and compress  DNN models with only minor performance degradation.
Specifically,  we propose to first directly factorize the weight matrix
using fixed-point (+1, 0 and -1) representation followed by recovering the 
(pseudo) full precision submatrices. We also propose an effective and practical technique
called weight balancing, which makes our fine-tuning (retraining) much more stable. 
We demonstrate the effects of the
direct fixed-point factorization, full precision weight recovery, weight balancing and 
whole-model performance of AlexNet \cite{krizhevsky2012imagenet},
VGG-16 \cite{simonyan2014very}, and ResNet-50 \cite{he2015deep} on ImageNet classification task. 
The main contributions of this paper can be summarized as follows:
\begin{itemize}
\item We propose the FFN framework based on direct fixed-point factorization 
for DNN acceleration
and compression, which is much more flexible and accurate.
\item Based on fixed point factorization, we propose a novel 
full precision weight recovery method, which makes it possible to 
make full use of the pretrained models even for very deep architectures 
like deep residual networks (ResNets) \cite{he2015deep}.
\item We investigate the weight imbalance problem generally existing in 
matrix/tensor decomposition based DNN acceleration methods. 
Inspired by weights initialization methods, we present an effective 
weight balancing technique to stabilize the fine-tuning stage of DNN models. 
\end{itemize}

\section{Related Work}

CNN acceleration and compression are widely studied in recent years. 
We mainly list works that are closely related with ours,
i.e., the matrix decomposition based methods and fixed-point quantization 
based methods. 

Deep neural networks are usually over-parameterized and the 
redundancy can be removed using  low-rank approximation
of filter matrix as shown in the work of \cite{denil2013predicting}. 
Since then, many low-rank based methods
have been proposed. Jaderberg \cite{jaderberg2014speeding} proposed to use
filter low-rank approximation and data reconstruction to lower the
approximation error.
Zhang et al. \cite{zhang2015accelerating} presented a novel nonlinear data 
reconstruction method, which allows asymmetric reconstruction to 
prevent error accumulation across layers. Their method achieved 
high speed-up on VGG-16 model with minor increase 
on top-5 error for ImageNet \cite{russakovsky2015imagenet} 
classification. 
Low-rank tensor decomposition methods like 
CP-decomposition \cite{lebedev2014speeding}, Tucker decomposition \cite{kim2015compression}
and Block Term Decomposition (BTD) \cite{wang2016accelerating} are also investigated 
and showed high speed-up and energy reduction.

Fixed-point quantization based methods are also investigated by 
several recent works. Soudry et al. developed the Expectation 
Backpropagation (EBP) \cite{cheng2015training} method, which is a variational Bayes method
to binarize both weights and neurons and achieved good results
for fully connected networks on MNIST dataset. In the work of 
BinaryConnect \cite{courbariaux2015binaryconnect}, 
the authors proposed to use binary weights for
forward and backward computation while keep a full-precision version
of weights for gradients accumulation. Good results have been
achieved on small datasets like MNIST, CIFAR-10 and SVHN. 
Binary-Weight-Network (BWN) and XNOR-net were proposed in a more recent work \cite{rastegari2016xnor}, which was 
among the first ones to evaluate the performance of binarization on large-scale
datasets like ImageNet \cite{russakovsky2015imagenet} and yielded good results. 
These methods train neural networks from scratch and can barely benefit from
pretrained networks. Hwang et al. \cite{hwang2014fixed} 
found a way by first quantize pretrained weights
using a reduced number of bits, followed by retraining. However, their method
achieved good results only for longer bits on small datasets and heavily relied on carefully choosing 
the step size of quantization using exhaustive search. 
The scalability on large-scale datasets remained unclear. 

Besides low-rank based and fixed-point quantization based 
methods mentioned above, there have been other approaches. 
Han et al. \cite{han2015learning} utilized network pruning to remove 
low-saliency parameters and small-weight connections to reduce parameter size. 
Product quantization was investigated in the work of \cite{wu2016quantized} to compress 
and speed-up DNNs at the same time. Teacher-student 
architectures \cite{hinton2015distilling, romero2014fitnets} were also well studied and
achieved promising results.

Unlike previous works, we explore fixed-point factorization on weight matrix.
It is nontrivial to utilize fixed-point factorization of weight matrices.
One may argue to use full precision matrix decomposition like SVD,
followed by fixed point quantization of the decomposed submatrices. However, this 
kind of method has an obvious shortcoming: the matrix approximation
is optimized for the full precision submatrices, but not for the fixed-point representation,
which is our main target. On the contrary, in our proposed FFN architecture, we directly
factorize weight matrix into fixed-point format in an end-to-end way.

\section{Approaches}

Our method exploits the weight matrix approximation method for deep
neural network acceleration and compression. Unlike many previous
low-rank based matrix decomposition methods which use floating point values
for the factorized submatrices, our method aims at fixed-point factorization
directly. 

To more efficiently make use of the pretrained weights, a novel pseudo 
full-precision weight matrix recovery method is introduced in addition to 
the direct fix-point factorization. Thus the information of the pretrained models
is divided into two parts: the first one is the fixed-point factorized 
submatrices and the second one resides in the pseudo full-precision weight
matrices, which on the other hand, will be transferred to the fixed-point weight matrices
during the fine-tuning stage. 

Moreover, we find that fine-tuning becomes much harder after decomposition, which
is also observed in the work of \cite{zhang2015accelerating}, i.e., a small
learning rate results in poor local optimum while a large learning rate may
discard the initialization information. Based on our empirical results and theoretical analysis,
we propose the weight balancing technique, which makes the fine-tuning
more efficient and has an important role in our whole framework. 

We will present our novel fixed-point factorization, pseudo full-precision 
weight recovery and weight balancing methods at length in section \ref{method:FPF}, 
\ref{method:FWR} and \ref{method:WB} respectively.

\subsection{Fixed-point Factorization of Weight Matrices}
\label{method:FPF}
A general deep neural network usually has multiple fully connected layers and / or 
convolutional layers. For the fully connected layers, the output signal vector
$s_o$ is computed as:
\begin{equation}
\label{forward}
s_o=\phi(Ws_i+b)
\end{equation}
where $s_i$ is the input signal vector and $W$ and $b$ are the weight matrix and the bias term respectively.
For a convolutional layer with $n$ filters of size $w\times{h}\times{c}$ where $w$, $h$ and $c$ 
are the kernel width and height and the number of input feature maps, if we reshape the kernel
and the input volume at every spatial positions, the feedforward pass of the convolution
can also be expressed by equation \ref{forward}. Thus our decomposition is conducted
on the weight matrix $W$.

In this subsection we propose to directly factorize the weight matrices
into fixed-point format. More specifically, in our framework, full precision weight matrix 
$W\in{R^{m\times{n}}}$ of a given pretrained model is approximated by a weighted sum of 
outer products of several ($k$) vector pairs with only 
ternary (+1, 0 and -1) entries, which is referred to as the semidiscrete decomposition (SDD)
in the following format: 
\begin{equation}
\label{SDD1}
\begin{aligned}
	& \underset{X, D, Y}{\text{minimize}}
	& & {\parallel{W-XDY^T}\parallel_F^2} \\
	= & \underset{\{x_i\}, \{d_i\}, \{y_i\}}{\text{minimize}}
	& & {\parallel{W-\sum_i^kd_ix_iy_i^T}\parallel_F^2} \\
\end{aligned}
\end{equation}
where $X\in{\{-1,0,+1\}^{m\times{k}}}$ and $Y\in{\{-1,0,+1\}^{n\times{k}}}$ and 
$D\in{R_+^{k\times{k}}}$ is a nonnegative diagonal matrix. Note that
throughout this paper, we utilize the symbol $k$ to represent the 
dimension of the SDD decomposition.

One advantage of fixed-point factorization method over direct 
fixed-point quantization is that there is much more room for 
us to control the approximation error. Consider the decomposition
on weight matrix $W\in{R^{m\times{n}}}$, we can choose different $k$
to approximate $W$ as accurate as possible. (Note that $k$ can be larger
than both $m$ and $n$). This also makes it possible to choose different
$k$ for different layers according to the redundancy of that layer. 
Thus our fixed-point decomposition method can be much more flexible
and accurate than direct quantization method.

\begin{algorithm}
\begin{algorithmic}[1]
\REQUIRE weight matrix $W\in{R^{m\times{n}}}$
\REQUIRE non-negative integer $k$
\ENSURE $X\in{\{+1,0,-1\}^{m\times{k}}}$
\ENSURE $Y\in{\{+1,0,-1\}^{n\times{k}}}$
\ENSURE diagonal matrix $D\in{R_+^{k\times{k}}}$
\STATE $d_i\gets 0$ for $i=1, \cdots, k$
\STATE Select $Y\in\{-1,0,1\}^{n\times k}$
\WHILE {not converge}
\FOR {$i=1, \cdots, k$}
\STATE $R\gets{W-\sum_{j\neq{i}}d_jx_jy_j^T}$
\STATE Set $y_i$ to the $i$-th column of $Y$
\WHILE {not converge}
\STATE compute $x_i\in\{-1,0,1\}^m$ given $y_i$ and $R$
\STATE compute $y_i\in\{-1,0,1\}^n$ given $x_i$ and $R$
\ENDWHILE
\STATE Set $d_i$ to the average of $R\circ{x_iy_i^T}$ over the non-zero locations of $x_iy_i^T$
\STATE Set $x_i$ as the $i$-th column of $X$ , $y_i$ the $i$-th column of $Y$ and $d_i$ the $i$-th diagonal value of $D$
\ENDFOR
\ENDWHILE
\end{algorithmic}
\caption{Improved SDD decomposition}
\label{alg:improved_sdd}
\end{algorithm}

Because of the ternary constraints in \ref{SDD1}, the computation of SDD is a NP-hard problem. 
Kolda and O'Leary \cite{Kolda1999A} proposed to obtain an approximate local solution by greedily 
finding the best next $d_ix_iy_i$. 
To further reduce the approximation error of the decomposition,  we refine 
their algorithm as in Algorithm \ref{alg:improved_sdd} by 
iteratively minimizing the residual error.

\begin{figure}[t]
\begin{center}
\includegraphics[scale=0.3]{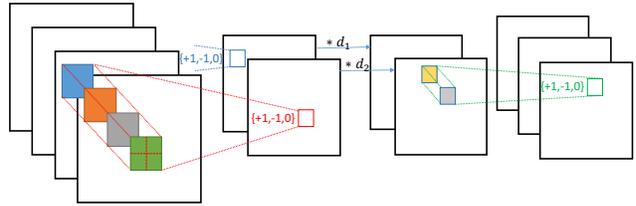}
\end{center}
   \caption{New layers used in our FFN architecture to replace the original convolutional layers.}
\label{fig:FFN}
\end{figure}

Once the decomposition is done, we can replace the original weights $W$ with
the factorized ones, i.e., the $X,Y$ and $D$. More formally, for convolutional
layers, the original layer is replaced by three layers: the first one is a convolutional 
layer with $k$ filters of size $w\times{h}\times{c}$, which are all with ternary values;
The second layer is a ``channel-wise scaling layer", i.e., each of the $k$ feature maps
is multiplied by a scaling factor; The last layer is another convolutional layer with
$n$ filters of size $1\times{1}\times{k}$, which also have ternary values.
Figure \ref{fig:FFN} illustrates the architecture of our new layers in FFN network.

\subsection{Full-precision Weight Recovery}
\label{method:FWR}

Our fixed-point factorization method is much more accurate than direct binarization
or ternarization method and many other fixed-point quantization methods. But there
is still the need of fine-tuning to restore the precision of DNN models. Like most of
current fixed-point quantization based accelerating methods, we want to use the 
quantized weights (the $X, Y$ in our case) during the forward and backward propagation while use the
full-precision weights for gradient accumulation. However, after factorization, 
the full-precision weights are lost, i.e., the original $W$ cannot be used for gradient 
accumulation any longer.  A simple solution is to use the floating point version 
of $X$ and $Y$ as full-precision weights to accumulate gradients. But this is far from 
satisfactory, as can be seen from section \ref{exp:FWR}.

In this subsection, we present our novel full-precision weight recovery method 
based on the pretrained weights to make the fine-tuning stage much easier. Our motivation is very simple, we 
recovery the full-precision version of $X$ and $Y$, indicated by
$\hat{X}$ and $\hat{Y}$, which can better approximate $W$.
Note that at the same time, we must make sure that $\hat{X}$ and 
$\hat{Y}$ will be quantized into $X$ and $Y$ in quantization stage.
We can treat our full-precision weight recovery method as an inversion of 
current fixed-point quantization methods. In fixed-point quantization based
DNN acceleration and compression methods, we quantize each element of 
the full-precision weight matrices into the nearest fixed-point format value.
While in our method, we have got the fixed-point version of the weights 
through fixed-point decomposition, and we need to
determine from which value the fixed-point element is quantized.
We turn this problem into an optimization problem as follows:
\begin{equation}
\label{recovery}
\begin{aligned}
	& \underset{\hat{X}, \hat{Y}}{\text{minimize}}
	& & {\parallel{W-\hat{X}D\hat{Y}^T}\parallel}_F^2 \\
	& \text{subject to}
	& & |\hat{X}_{ij}-X_{ij}|<0.5, \forall{i,j} \\
    & & & |\hat{Y}_{ij}-Y_{ij}|<0.5,\forall{i,j}
\end{aligned}
\end{equation}

Here the two constraints are introduced to ensure that the $\hat{X}$ and $\hat{Y}$ 
will be quantized to $X$ and $Y$. 
The problem can be efficiently solved by alternative method.
Note we constraint $\hat{X}$ and $\hat{Y}$ to be always between -1.5 and 1.5 to 
alleviate overfitting and during fine-tuning, we also clip the weights within [-1.5, 1.5] interval as well.

During fine-tuning stage, we quantize the full-precision weights of $\hat{X}$ and $\hat{Y}$ 
according to the following equation (before weight balancing described in the next subsection):
\begin{equation}
\label{quantize}
q(A_{ij})=\left\{
	\begin{array}{ccl}
	{+1} & 0.5<A_{ij}<1.5 \\
	{0}   & -0.5\le{A_{ij}}\le{0.5} \\
	{-1}  & -1.5<A_{ij}<-0.5
	\end{array}
\right.
\end{equation}
The quantized weights are used to conduct forward and backward computation and 
the full-precision weights $\hat{X}$ and $\hat{Y}$ are used to accumulate gradients. 
Both $X$ and $Y$ will change during fine-tuning because of
the updates of $\hat{X}$ and $\hat{Y}$, for example, 
some elements of $X$ and $Y$ will turn from 0 to 1 and so on.
We argue that, for example, both 0.499 and 0.001 will be quantized to 0 according to 
Equation \ref{quantize}. But at fine-tuning stage, 0.499 has higher probability than 0.001 to 
turn to 1. And this kind of  information resides in the full-precision weight
matrices and is transferred to the quantized weights during fine-tuning. Note that the 
full-precision weights won't be retained after fine-tuning, and there 
are only the quantized weights $X$ and $Y$ for prediction.

\subsection{Weight Balancing}
\label{method:WB}

\begin{figure*}
\centering
\subfigure[Forward and backward propagation ]{
\includegraphics[scale=0.45]{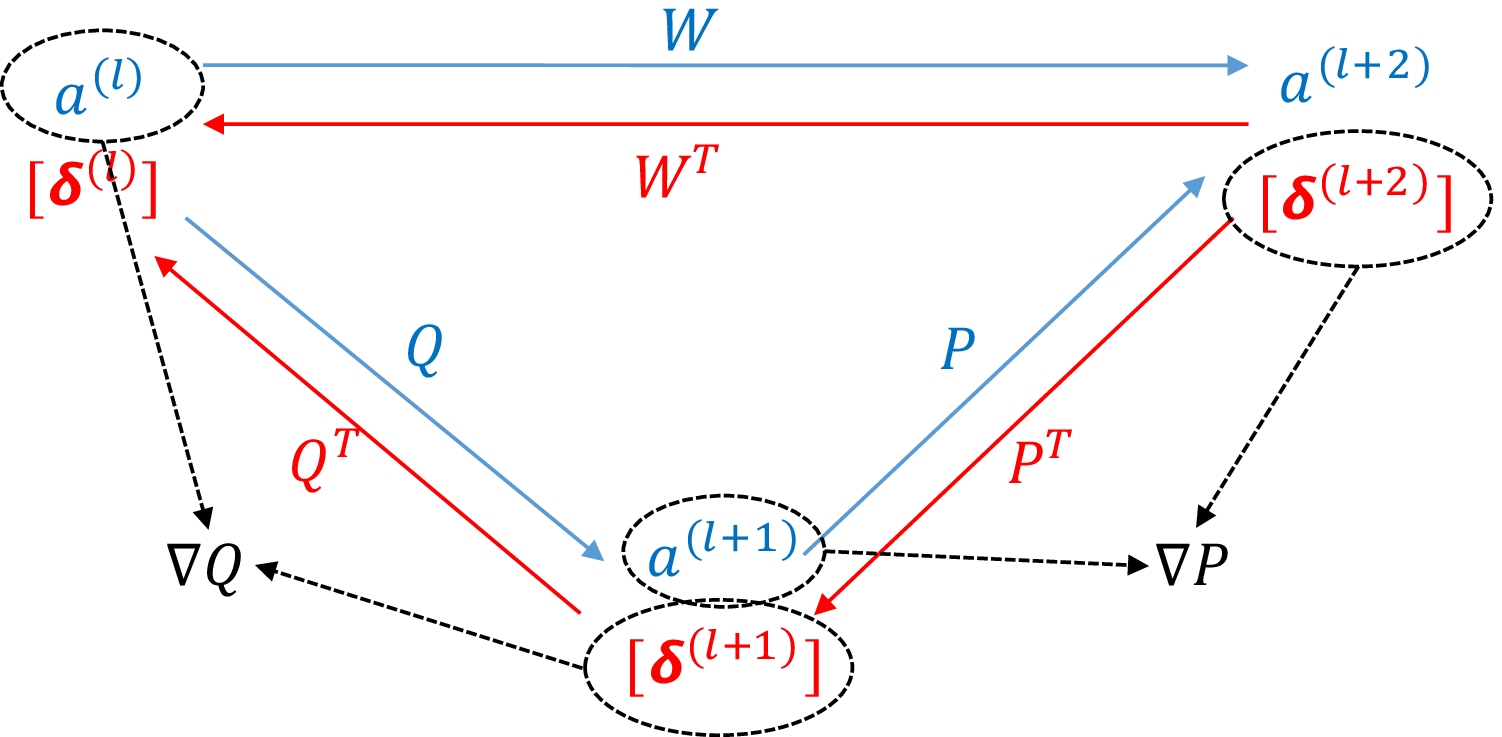}
\label{fig:BP:a}
}
\subfigure[Propagation after setting $Q'=\alpha*Q$ and $P'=P/\alpha$]{
\includegraphics[scale=0.45]{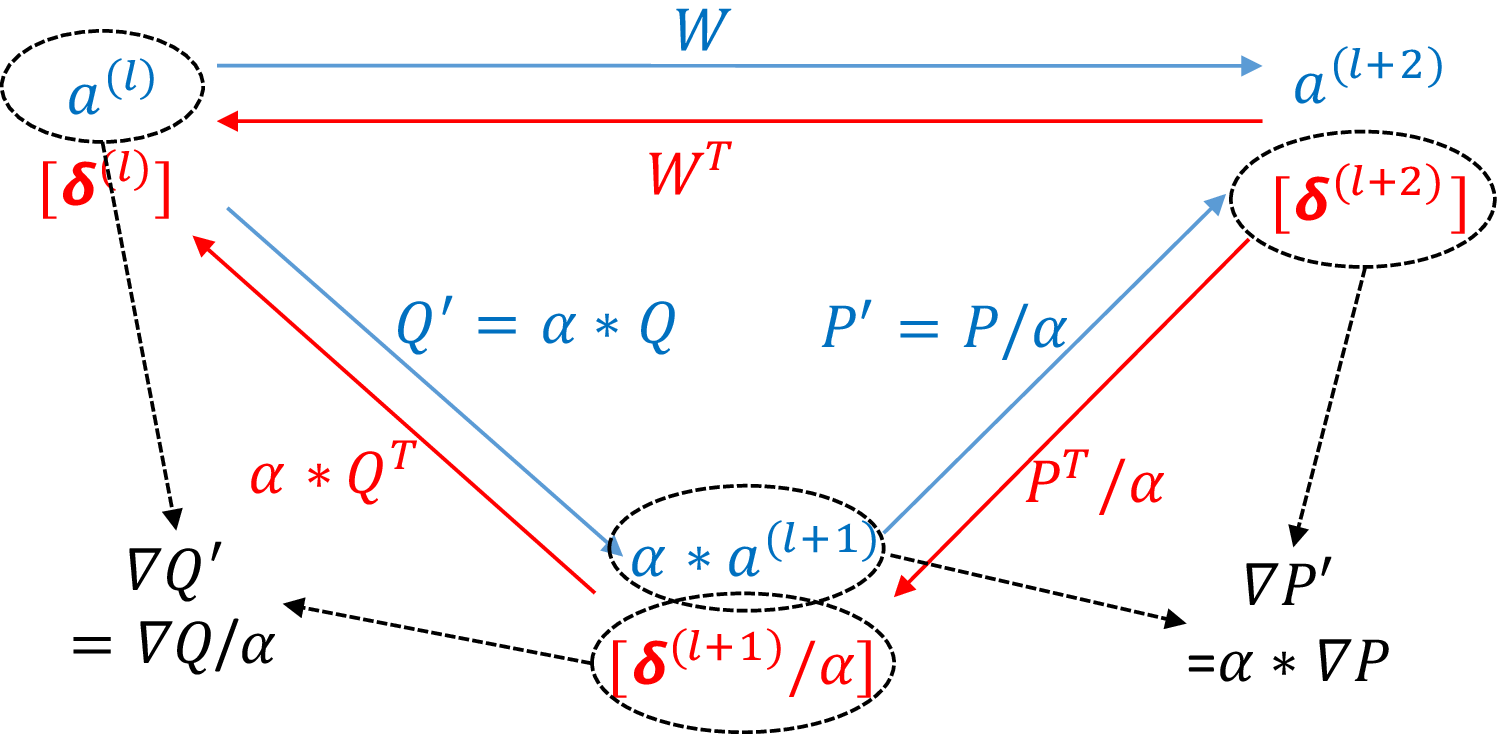}\
\label{fig:BP:b}
}
	\caption{Illustration of the cause of weight imbalance problem existing
in decomposition based methods.}
\label{fig:BP}
\end{figure*}

So far, we have presented our fixed-point decomposition and full-precision weight
recovery to improve the test-phase efficiency of deep neural networks. 
However, there is still a problem to be considered, which we refer to as 
weight imbalance.

Weight imbalance is a common problem of decomposition based methods, not
just existing in our framework (as also noticed in \cite{zhang2015accelerating}). This problem is caused by the non-uniqueness of
the decomposition. 

Considering a $L$ layers neural network, the forward computation is in the following format:
\begin{equation}
\left.
	\begin{array}{l}
		z^{(l+1)} = W^{(l)}a^{(l)}+b^{(l)} \\
		a^{(l+1)}=\phi(z^{(l+1)})
	\end{array}
\right.
\end{equation}
During back-propagation, the error term and the gradients of weights 
for each layer are as follows:
\begin{equation}
\label{error_term}
\delta^{(l)}=((W^{(l)})^T\delta^{(l+1)})\bullet\phi'(z^{(l)})
\end{equation}
\begin{equation}
\label{delta}
\nabla_{W^{(l)}}=\delta^{(l+1)}(a^{(l)})^T
\end{equation}
Here the ``$\bullet$" denotes the element-wise product operator. 
Note that for layer $l$, the inputs, outputs and the error term are represented
as $a^{(l)}$, $a^{(l+1)}$ and $\delta^{(l)}$.
From Equation \ref{delta} we can see that the gradients 
$\nabla_{W{(l)}}$ is proportional to this layer's
input $a^{(l)}$ and the next layer's error term $\delta^{(l+1)}$. While from
Equation \ref{error_term} we can see that the next layer's
error term $\delta^{(l+1)}$ is proportional to the next layer's weights $W^{(l+1)}$.

Suppose we have a weight matrix $W$, which is factorized 
into the product of two matrices $W=PQ$, i.e., the original layer with parameter 
$W$ is replaced by two layers with parameter $Q$ and $P$, as shown in 
Figure \ref{fig:BP:a}. 
If let $P'=P/\alpha$ and $Q'=\alpha*Q$, the decomposition becomes 
$W=P'Q'$ as shown in Figure \ref{fig:BP:b}. 
Figure \ref{fig:BP} shows that $Q$ has been enlarged by $\alpha$-times while
the gradients have become $1/\alpha$ of the original. And what happened to $P$ 
is opposite to $Q$. The consequence (suppose $\alpha\gg1$) is that during back-propagation, 
$P$ changes frequently while $Q$ almost stays untouched. At this time, 
one has to search for different learning rates for each layer. However, finding appropriate 
learning rates for every layer is 
quite a hard job especially for very deep neural networks.

In our framework, the weight matrix $W\in{R^{m\times n}}$ is replaced by
 $\hat{X}D\hat{Y}^T$, where the $\hat X\in R^{m\times k}$ and 
 $\hat Y \in R^{n\times k}$ is in the range of [-1.5, 1.5] while $D$ is at the scale of
about 0.00001 to 0.01. And for convolutional layers, $\hat{X}$ usually has much
more elements than $\hat Y$ because of the $w\times{h}$ spatial size of filters in $\hat{X}$.
To balance the weights into their appropriate scales, and inspired by the normalized
 weight initialization method proposed in \cite{Glorot2010Understanding}, 
 we develop the following weight balancing approaches:
 
 First, we want to find the scaling factor $\lambda_X$ and $\lambda_Y$
 for $\hat X$ and $\hat Y$,
 which are proportional to the square root of the sum of the number of their rows and columns. 

Second, we try to make the balanced ${D}$ close to identity matrix by
setting the mean value of the elements along the diagonal to one. Because for
fully-connected layer, $D$ is a element-wise scaling factor and for convolutional layers,
$D$ is a channel-wise scaling factor. Thus making $D$ close to one will not
affect the calculation of gradients much.

This can be expressed by the equation \ref{balance} where $\tilde{X}$, $\tilde{Y}$ and $\tilde{D}$
represent the balanced version of weight matrices. And the $\varphi$ is introduced to make sure
that the scaling factor $\lambda_X$ and $\lambda_Y$ are proportional to the square root of
the sum number of rows and columns.
\begin{equation}
\label{balance}
\left\{
	\begin{array}{l}
		 \tilde{X}=\lambda_X*\hat{X}=\frac{\varphi}{\sqrt{m+k}}*\hat{X} \\
		 \tilde{Y}=\lambda_Y*\hat{Y}=\frac{\varphi}{\sqrt{n+k}}*\hat{Y} \\
		 \tilde{D}=\frac{D}{\lambda_X*\lambda_Y} \\
		 \text{mean}(\tilde{D})=1
	\end{array}
\right.
\end{equation}

Once we have got the scaling factors of $\lambda_X, \lambda_D$ and $\lambda_Y$, 
we can use the balanced weights $\tilde{X}, \tilde{D}$ and $\tilde{Y}$ during 
back-propagation. Note that we also need to scale the quantization function accordingly
in the following form where $\lambda$ can be  $\lambda_X$ and $\lambda_Y$ respectively:
\begin{equation}
\label{quantize_scaled}
q(A_{ij})=\left\{
	\begin{array}{ccl}
	{+1*\lambda} & 0.5*\lambda<A_{ij}<1.5*\lambda \\
	{0}   & -0.5*\lambda\le{A_{ij}}\le{0.5}*\lambda \\
	{-1*\lambda}  & -1.5*\lambda<A_{ij}<-0.5*\lambda
	\end{array}
\right.
\end{equation}

\subsection{Fine-tuning}
Thanks to the full-precision weight recovery strategy and
weight balancing method proposed in this paper, we can 
easily fine-tune the factorized network to restore accuracy.
Specifically, we keep the balanced pseudo full-precision 
weight matrices ($\tilde{X}$ and $\tilde{Y}$) as reference.
During fine-tuning stage, we quantize 
$\tilde{X}$ and $\tilde{Y}$ according to equation \ref{quantize_scaled} 
and the quantized weights are used in the
 forward and backward computation. While the gradients 
 are accumulated by the full-precision weights, i.e., 
 $\tilde{X}$ and $\tilde{Y}$, to make improvements.
The full-precision weight recovery and weight balancing are 
introduced to facilitate convergence of the fine-tuning
stage. However, at test time, we only need the fixed point
$X$, $Y$ and the diagonal floating-point $D$ for prediction.

\subsection{Complexity Analysis}
In this section, we will analyze the computing complexity of our framework
for convolutional layers, which dominates the operations of convolutional neural networks. 
Fully-connected layers can be analyzed in a similar way.

For convolutional layers, the width and height of output feature maps 
are denoted as $W'$ and $H'$. Considering convolution with kernel
of size $w\times{h}\times{c}\times{n}$, the computation of the original
layer is given by:
\begin{equation}
C_{mul}=C_{add}=W'*H'*(w*h*c*n)
\end{equation}
In our FFN architecture, the computation turns to be:
\begin{equation}
\label{complexity}
\begin{aligned}
	C_{mul}&=W'*H'*k\\
	C_{add}&=(1-\alpha)*W'*H'*(w*h*c+n)*k\\
	&\approx(1-\alpha)*W'*H'*(w*h*c*n)
\end{aligned}
\end{equation}
Here, the $\alpha$ denotes the sparsity of weight matrix for this layer.
For a common convolutional layer, the $w*h*c*n$ is usually thousands of
times of $k$, thus the number of multiply operation can be dramatically reduced.
The $c$, $n$ and $k$ are usually at the same scale, making the addition 
operation about $(1-\alpha)$ times of the original. In our experiments, we
find that $\alpha$ is around 0.5. Thus our method only requires about
half of operations compared to that using binary weights. We refer to 
section \ref{exp:efficiency_analysis} for more detail.

\section{Experiments}
In this section, we comprehensively evaluate our method on ILSVRC-12 
\cite{russakovsky2015imagenet} 
image classification benchmark, which has 1.2M training examples and 
50K validation examples. We firstly examine the effects of each individual 
component in FFN, i.e., fixed-point factorization, full-precision weight 
recovery, and weight balancing. The whole-model 
ILSVRC-12 \cite{russakovsky2015imagenet} classification 
performance is also evaluated based on AlexNet \cite{krizhevsky2012imagenet},
VGG-16 \cite{simonyan2014very}, and ResNet-50 \cite{he2015deep}, 
demonstrating the effectiveness of our FFN framework.

\subsection{Effectiveness of Each Part}

In this subsection we thoroughly analyze the effectiveness of each part
in our unified FFN framework. 

\subsubsection{Fixed-point Factorization}
\label{exp:FPF}
In theory, our method can approximate weight matrix $W$ as accurate
as possible by choosing large $k$, i.e., the dimension of SDD decomposition. 
We can also utilize different
$k$ for different layers. Thus our method can be much more
accurate and flexible than the direct fixed-point quantization.
In this section, we evaluate the weight matrix approximation error and
classification accuracy under different $k$. 

We use the second convolutional layer of AlexNet for demonstration, which
is the most time-consuming layer during the test phase. There are two groups
in this layer, each is of size $5\times{5}\times{48}\times{128}$.
We choose the same $k$ for these two groups and evaluate the average 
of weight matrix approximation error. Here, weight matrix approximate error
is defined as:
\begin{equation}
r=\frac{\parallel{W-XDY^T}\parallel_F^2}{\parallel{W}\parallel_F^2}
\end{equation}

\begin{figure}[t]
\centering
\includegraphics[scale=0.5]{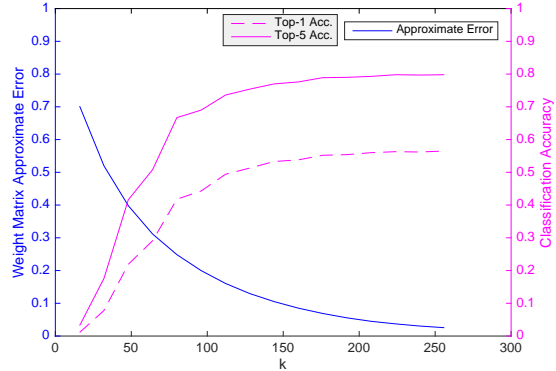}
   \caption{Weight approximation error and classification accuracy on ImageNet when choosing
   different $k$ for the second convolutional layer of AlexNet.}
\label{fig:err_acc}
\end{figure}

Figure \ref{fig:err_acc} illustrates the approximation error and 
the accuracy on ImageNet classification task. From Figure \ref{fig:err_acc}, 
we can see that as $k$ increases, the approximation error tends
to zero and the accuracy stays closer to the original AlexNet. 

The classification accuracy after all layers are processed is given in the second
row of Table \ref{modular_results_alexnet} (denoted as FFN-SDD), 
demonstrating that our fixed-point factorized method can produce a
good initialization.

\begin{table} \footnotesize
\caption{Results of different settings on AlexNet .}
\centering
 \begin{tabular}{|l|c|c|}
 \hline
  Model  & Top-1 Acc. (\%) & Top-5 Acc. (\%)\\
 \hline
  AlexNet \cite{krizhevsky2012imagenet} & 57.1 & 80.2  \\
 \hline
FFN-SDD & 32.9 & 57.0   \\
 \hline
FFN-Recovered & 57.0 & 80.1   \\
\hline
FFN-W/O-FWR & 53.4 & 77.2   \\
\hline
FFN-W/O-WB & 51.9 & 76.6   \\
\hline
FFN & 55.5 & 79.0   \\
 \hline
 \end{tabular}
\label{modular_results_alexnet}
\end{table}

\begin{figure*}[t]
\centering
\includegraphics[scale=0.38]{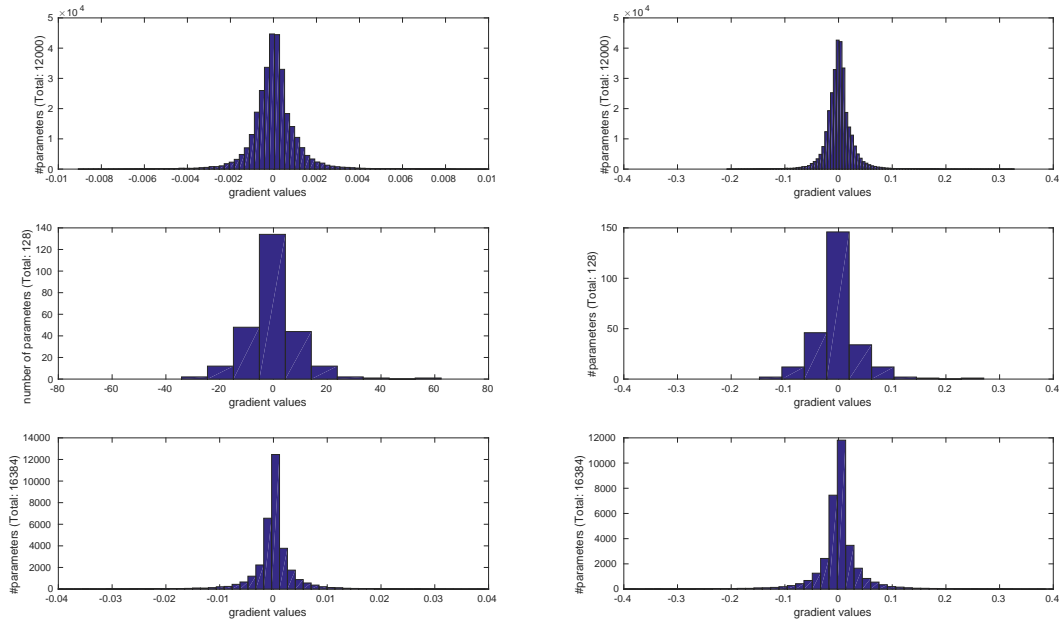}
   \caption{Gradient distribution of the second convolutional layer of AlexNet
   before (first column) and after (second column) weight balancing. Three rows
   correspond to $X$, $D$ and $Y$ respectively.}
\label{fig:diag_grad}
\end{figure*}

\subsubsection{Full-precision Weight Recovery}
\label{exp:FWR}
In this subsection,  we evaluate the effect of our full-precision weight recovery
method. During the fine-tuning stage, gradients are accumulated by the full-precision
weights, thus the initial values may affect the evolution of learning process.
To show that the pseudo full-precision weights recovered by our method can 
actually represent the original weights, we evaluate the performance of  
recovered weights on AlexNet. The results are given in the third row of 
Table \ref{modular_results_alexnet} (FFN-Recovered). Both the top-1 and top-5 
classification accuracy are very close to the original AlexNet model.

To further demonstrate the effectiveness of the weight recovery strategy, we also
compare with FFN model without full-precision weight recovery (FFN-W/O-FWR, 
weight balancing method is incorporated)
in Table \ref{modular_results_alexnet}. Without full-precision weight recovery, 
the top-5 accuracy decreases 1.8\% compared to FFN.

\subsubsection{Weight Balancing}
\label{exp:WB}

Weight balancing is introduced to make the fine-tuning stage more reliable. 
In Table \ref{modular_results_alexnet}, we report the best results achieved
 without weight balancing (FFN-W/O-WB) 
compared with that of using weight balancing (FFN) on AlexNet.
The weight balancing scheme greatly helps the fine-tuning stage, leading 
to 3.6\%/2.4\% improvement in the top-1/top-5 classification accuracy.

To further illustrate the gradients imbalance problem as well as to show the 
effectiveness of our novel weight balancing method, we extract the gradients
of the second convolutional layer of AlexNet, as shown in Figure \ref{fig:diag_grad}.
The left and right columns represent the gradient distribution before and
after applying our weight balancing method. From Figure \ref{fig:diag_grad},
we discover that after decomposition, the gradients of the three new layers
differ significantly from each other, while after weight balancing, most gradients
lie within the interval [-0.1, 0.1] for all layers. Using weight balancing method allows to use
the same learning rate for all layers, which is very important for fine-tuning,
especially for very deep networks.

\subsection{Whole-model Performance on ILSVRC-12}
In this subsection, we evaluate the performance of our FFN on ImageNet 
classification task. We report  top-1 and top-5 accuracy using the 
224$\times$224 center crop.
Experiments are conducted on three commonly used
CNN models, i.e., AlexNet \cite{krizhevsky2012imagenet}, 
VGG-16 \cite{simonyan2014very} and ResNet-50 \cite{he2015deep}. All of these models are
downloaded from Berkeley's \emph{Caffe model zoo} \cite{jia2014caffe} without any change and are
also used as baselines for comparison. Our accelerating strategy is to approximate
the original weight matrices using the proposed fixed-point decomposition, 
full-precision weight recovery and weight balancing method. 
After that, fine-tuning (retraining) the whole network for the ImageNet classification
task is needed to retain accuracy.

\subsubsection{AlexNet}

Alexnet was proposed in  \cite{krizhevsky2012imagenet} and was the winner of
ILSVRC 2012 \cite{russakovsky2015imagenet} image classification task. 
This network has 61M parameters and more than 
95\% of them reside in the fully-connected layers. Thus we choose relatively 
smaller decomposition dimension $k$ for fully-connected layers for a higher compression rate. 
Specifically, for the convolutional layers
with 4-D weights of size $w\times h \times c \times n$, we choose decomposition dimension 
$k=\text{min}(w*h*c, n)$. And for the last three fully-connected layers, $k$ is set to 2048, 3072 and 1000 respectively.
The resulting architecture has 60M parameters, of which mostly are -1, 0, or 1.
At fine-tuning stage, images are resized to $256\times 256$ pixel size as the
same with original Alexnet. 

We also compare our method with the following approaches, whose results
on ImageNet dataset are publicly available. Note that the BWN \cite{rastegari2016xnor} method only report their
results on AlexNet with batch normalization \cite{ioffe2015batch}, so in order to compare with 
their results, we also report our results using batch normalization with the 
same settings as in \cite{rastegari2016xnor}.
\setlength{\pltopsep}{6pt}
\begin{compactitem}
\item \textbf{BWN}: \cite{rastegari2016xnor}: Binary-weight-network, using binary weights and floating point scaling factors;
\item \textbf{BC}: \cite{courbariaux2015binaryconnect}: BinaryConnect, using binary weights, reported by \cite{rastegari2016xnor};
\item \textbf{LDR} \cite{miyashita2016convolutional}: Logarithmic Data Representation, 4-bit logarithmic activation and 5-bit logarithmic weights.
\end{compactitem}
The results are listed in Table \ref{results_alexnet}. 
The suffix $BN$ indicates that batch normalization \cite{ioffe2015batch} is used.
From the results, we can see that without batch normalization, our method 
only has a 1.2\% drop in the top-5 accuracy. Our method can outperform
the best results by 2.2 percentages on top-5 accuracy if batch normalization
is incorporated.

\begin{table} \footnotesize
\caption{Comparison on AlexNet (Suffix BN indicates using batch normalization \cite{ioffe2015batch}).}
\centering
 \begin{tabular}{|l|c|c|}
 \hline
  Model  & Top-1 Acc. (\%) & Top-5 Acc. (\%)\\
 \hline
  AlexNet \cite{krizhevsky2012imagenet} & 57.1 & 80.2  \\
 \hline
  AlexNet-BN \cite{simon2016imagenet} & 60.1 & 81.9  \\
 \hline
BC-BN \cite{courbariaux2015binaryconnect} & 35.4  & 61.0   \\
 \hline
BWN-BN \cite{rastegari2016xnor} & 56.8  & 79.4   \\
 \hline
LDR \cite{miyashita2016convolutional} & -  & 75.1   \\
 \hline
FFN & 55.5 & 79.0   \\
 \hline
FFN-BN & 59.1  & 81.6   \\
 \hline
 \end{tabular}
\label{results_alexnet}
\end{table}

\subsubsection{VGG-16}

VGG-16 \cite{simonyan2014very} uses much wider and deeper structure than AlexNet, 
with 13 convolutional layers and 3 fully-connected layers. 
We use the same rules to choose the decomposition dimension $k$ and we set $k=3138, 3072, 1000$ for three 
fully-connected layers respectively, resulting in approximately the same number of 
parameters as the original VGG-16 model.
During fine-tuning, we resize images to 256 pixels at the smaller dimension.

The results are illustrated in Table \ref{results_vgg16}. 
We can see that after quntization, our method
even outperform the original VGG-16 model by 0.2\% on top-5 accuracy.

\begin{table} \footnotesize
\caption{Comparison on VGG-16.}
\centering
 \begin{tabular}{|l|c|c|}
 \hline
  Model  & Top-1 Acc. (\%) & Top-5 Acc. (\%)\\
 \hline
  VGG-16 \cite{simonyan2014very} & 71.1 & 89.9  \\
 \hline
LDR \cite{miyashita2016convolutional} & - & 89.0   \\
 \hline
FFN & 70.8 & 90.1   \\
 \hline
 \end{tabular}
\label{results_vgg16}
\end{table}

\subsubsection{ResNet-50}
To further evaluate the effectiveness of our FFN framework, we also conduct
experiments on the more challenging deep neural network, i.e., ResNet-50.
Residual Networks (ResNets) were proposed in \cite{he2015deep} which 
won the 1st place in the ILSVRC 2015 classification, 
detection and localization tasks. For simplicity, we choose the 50-layer 
architecture, which is the smallest ResNets that outperforms all previous models. 

The ResNet-50 architecture has a global average pooling layer before the 1000-way
fully-connected layer, thus the fully-connected layer has much fewer parameters than that in 
AlexNet \cite{krizhevsky2012imagenet} and VGG-16 \cite{simonyan2014very}.
To make the number of parameters the same as the original ResNet-50, we have to
choose relatively smaller $k$ for all convolutional layers.
Specifically, for a convolutional layer with kernel
size $w\times{h}\times{c}\times{n}$, we set $k=\frac{(w*h*c)*n}{w*h*c+n}$, i.e., 
for each layer, we keep the same number of parameters as the original layer. 
Even though, our method can still achieve promising performance, i.e., with
1.3\% drop in the top-5 accuracy as is shown in Table \ref{results_resnet50}.
Choosing higher $k$ for convolutional layers as is done for AlexNet and 
VGG-16 may further reduce the classification error.

\begin{table} \footnotesize
\caption{Comparison on ResNet-50.}
\centering
 \begin{tabular}{|l|c|c|}
 \hline
  Model  & Top-1 Acc. (\%) & Top-5 Acc. (\%)\\
 \hline
  ResNet-50 \cite{he2015deep} & 75.2 & 92.2  \\
 \hline
FFN & 72.7 & 90.9   \\
 \hline
 \end{tabular}
\label{results_resnet50}
\end{table}

\subsection{Efficiency Analysis}
\label{exp:efficiency_analysis}
In this section, the computational complexity and storage requirement of the 
proposed FFN are analyzed and compared to the original networks and 
networks using binary weights. Our architecture use ternary
weights, and we empirically find that about a half of weights are zeros. Thus 
the computational complexity is about a half of binary based method like BC \cite{courbariaux2015binaryconnect}.
The disadvantage of using ternary weights is that it needs a little more storage
than binary weights. Specifically, our ternary method has about 1.5-bit weight 
representation, because of the sparsity. Table \ref{complexity} shows 
the computation and storage on AlexNet, VGG-16 and ResNet-50 in detail.

\begin{table} \footnotesize
\caption{Operations and storage requirements. Mul and Add represent the number of 
multiply and addition operation. Bytes indicates the number of byte needed
to store the weights. 
All numbers are accounted for convolutional layers and fully-connected layers.
}
\centering
\begin{tabular}{|c|c|c|c|c|}
\hline
\multicolumn{2}{|c|}{Model}  &  AlexNet  &  VGG-16  &  ResNet-50  \\
\hline
\hline
\multirow{3}{*}{Original} & Mul & 725M & 15471M & 4212M \\
\cline{2-5}
  & Add & 725M & 15471M & 4212M \\
\cline{2-5}
  & Bytes & 244M & 528M & 97.3M \\
\hline
\hline
\multirow{3}{*}{Binary} & Mul & 0.66M & 13.5M & 10.6M \\
\cline{2-5}
  & Add & 725M & 15471M & 4212M \\
\cline{2-5}
  & Bytes & 7.7M & 16.6M & 3.1M \\
\hline
\hline
\multirow{3}{*}{FFN} & Mul & 0.66M & 11.7M & 4.4M \\
\cline{2-5}
  & Add & 392M & 8631M & 1907M \\
\cline{2-5}
  & Bytes & 11.5M & 25.8M & 4.9M \\
\hline
\end{tabular}
\label{complexity}
\end{table}

\section{Conclusion}
We introduce a novel fixed-point factorized framework, named FFN, for deep neural networks
acceleration and compression. To make full use of the pretrained models, 
we propose a novel full-precision weight recovery method, which makes 
the fine-tuning more efficient and effective. Moreover, we present a 
weight balancing technique to stabilize fine-tuning stage. 
Extensive experiments on AlexNet, VGG-16 and ResNet-50 show that the
proposed FFN only requires one-thousandth of multiply operations with comparable accuracy.

\paragraph*{Acknowledgement.}
This work was supported in part by National Natural Science Foundation of China (No.61332016) 
and the Scientific Research Key Program of Beijing Municipal Commission of Education (KZ201610005012).

{\small
\bibliographystyle{ieee}
\bibliography{mybibfile}

\begin{thebibliography}{10}\itemsep=-1pt

\bibitem{cheng2015training}
Z.~Cheng, D.~Soudry, Z.~Mao, and Z.~Lan.
\newblock Training binary multilayer neural networks for image classification
  using expectation backpropagation.
\newblock {\em arXiv preprint arXiv:1503.03562}, 2015.

\bibitem{courbariaux2016binarynet}
M.~Courbariaux and Y.~Bengio.
\newblock Binarynet: Training deep neural networks with weights and activations
  constrained to+ 1 or-1.
\newblock {\em arXiv preprint arXiv:1602.02830}, 2016.

\bibitem{courbariaux2014low}
M.~Courbariaux, Y.~Bengio, and J.-P. David.
\newblock Low precision arithmetic for deep learning.
\newblock {\em arXiv preprint arXiv:1412.7024}, 2014.

\bibitem{courbariaux2015binaryconnect}
M.~Courbariaux, Y.~Bengio, and J.-P. David.
\newblock Binaryconnect: Training deep neural networks with binary weights
  during propagations.
\newblock In {\em Advances in Neural Information Processing Systems}, pages
  3123--3131, 2015.

\bibitem{denil2013predicting}
M.~Denil, B.~Shakibi, L.~Dinh, N.~de~Freitas, et~al.
\newblock Predicting parameters in deep learning.
\newblock In {\em Advances in Neural Information Processing Systems}, pages
  2148--2156, 2013.

\bibitem{dettmers20158}
T.~Dettmers.
\newblock 8-bit approximations for parallelism in deep learning.
\newblock {\em arXiv preprint arXiv:1511.04561}, 2015.

\bibitem{Glorot2010Understanding}
X.~Glorot and Y.~Bengio.
\newblock Understanding the difficulty of training deep feedforward neural
  networks.
\newblock {\em Journal of Machine Learning Research}, 9:249--256, 2010.

\bibitem{gupta2015deep}
S.~Gupta, A.~Agrawal, K.~Gopalakrishnan, and P.~Narayanan.
\newblock Deep learning with limited numerical precision.
\newblock {\em CoRR, abs/1502.02551}, 392, 2015.

\bibitem{han2015learning}
S.~Han, J.~Pool, J.~Tran, and W.~Dally.
\newblock Learning both weights and connections for efficient neural network.
\newblock In {\em Advances in Neural Information Processing Systems}, pages
  1135--1143, 2015.

\bibitem{he2015deep}
K.~He, X.~Zhang, S.~Ren, and J.~Sun.
\newblock Deep residual learning for image recognition.
\newblock {\em IEEE Conference on Computer Vision and Pattern Recognition
  (CVPR)}, 2016.

\bibitem{hinton2015distilling}
G.~Hinton, O.~Vinyals, and J.~Dean.
\newblock Distilling the knowledge in a neural network.
\newblock {\em arXiv preprint arXiv:1503.02531}, 2015.

\bibitem{hubara2016binarized}
I.~Hubara, D.~Soudry, and R.~E. Yaniv.
\newblock Binarized neural networks.
\newblock {\em arXiv preprint arXiv:1602.02505}, 2016.

\bibitem{hwang2014fixed}
K.~Hwang and W.~Sung.
\newblock Fixed-point feedforward deep neural network design using weights+ 1,
  0, and- 1.
\newblock In {\em 2014 IEEE Workshop on Signal Processing Systems (SiPS)},
  pages 1--6. IEEE, 2014.

\bibitem{ioffe2015batch}
S.~Ioffe and C.~Szegedy.
\newblock Batch normalization: Accelerating deep network training by reducing
  internal covariate shift.
\newblock {\em arXiv preprint arXiv:1502.03167}, 2015.

\bibitem{jaderberg2014speeding}
M.~Jaderberg, A.~Vedaldi, and A.~Zisserman.
\newblock Speeding up convolutional neural networks with low rank expansions.
\newblock {\em arXiv preprint arXiv:1405.3866}, 2014.

\bibitem{jia2014caffe}
Y.~Jia, E.~Shelhamer, J.~Donahue, S.~Karayev, J.~Long, R.~Girshick,
  S.~Guadarrama, and T.~Darrell.
\newblock Caffe: Convolutional architecture for fast feature embedding.
\newblock In {\em Proceedings of the ACM International Conference on
  Multimedia}, pages 675--678. ACM, 2014.

\bibitem{kim2016bitwise}
M.~Kim and P.~Smaragdis.
\newblock Bitwise neural networks.
\newblock {\em arXiv preprint arXiv:1601.06071}, 2016.

\bibitem{kim2015compression}
Y.-D. Kim, E.~Park, S.~Yoo, T.~Choi, L.~Yang, and D.~Shin.
\newblock Compression of deep convolutional neural networks for fast and low
  power mobile applications.
\newblock {\em arXiv preprint arXiv:1511.06530}, 2015.

\bibitem{Kolda1999A}
T.~G. Kolda and D.~P. O'Leary.
\newblock A semidiscrete matrix decomposition for latent semantic indexing in
  information retrieval.
\newblock {\em Acm Transactions on Information Systems}, 16(4):322--346, 1999.

\bibitem{krizhevsky2012imagenet}
A.~Krizhevsky, I.~Sutskever, and G.~E. Hinton.
\newblock Imagenet classification with deep convolutional neural networks.
\newblock In {\em Advances in neural information processing systems}, pages
  1097--1105, 2012.

\bibitem{lebedev2014speeding}
V.~Lebedev, Y.~Ganin, M.~Rakhuba, I.~Oseledets, and V.~Lempitsky.
\newblock Speeding-up convolutional neural networks using fine-tuned
  cp-decomposition.
\newblock {\em arXiv preprint arXiv:1412.6553}, 2014.

\bibitem{lin2015fixed}
D.~D. Lin, S.~S. Talathi, and V.~S. Annapureddy.
\newblock Fixed point quantization of deep convolutional networks.
\newblock {\em arXiv preprint arXiv:1511.06393}, 2015.

\bibitem{lin2015neural}
Z.~Lin, M.~Courbariaux, R.~Memisevic, and Y.~Bengio.
\newblock Neural networks with few multiplications.
\newblock {\em arXiv preprint arXiv:1510.03009}, 2015.

\bibitem{miyashita2016convolutional}
D.~Miyashita, E.~H. Lee, and B.~Murmann.
\newblock Convolutional neural networks using logarithmic data representation.
\newblock {\em arXiv preprint arXiv:1603.01025}, 2016.

\bibitem{rastegari2016xnor}
M.~Rastegari, V.~Ordonez, J.~Redmon, and A.~Farhadi.
\newblock Xnor-net: Imagenet classification using binary convolutional neural
  networks.
\newblock In {\em ECCV (4)}, volume 9908, pages 525--542. Springer, 2016.

\bibitem{romero2014fitnets}
A.~Romero, N.~Ballas, S.~E. Kahou, A.~Chassang, C.~Gatta, and Y.~Bengio.
\newblock Fitnets: Hints for thin deep nets.
\newblock {\em arXiv preprint arXiv:1412.6550}, 2014.

\bibitem{russakovsky2015imagenet}
O.~Russakovsky, J.~Deng, H.~Su, J.~Krause, S.~Satheesh, S.~Ma, Z.~Huang,
  A.~Karpathy, A.~Khosla, M.~Bernstein, et~al.
\newblock Imagenet large scale visual recognition challenge.
\newblock {\em International Journal of Computer Vision}, 115(3):211--252,
  2015.

\bibitem{simon2016imagenet}
M.~Simon, E.~Rodner, and J.~Denzler.
\newblock Imagenet pre-trained models with batch normalization.
\newblock {\em arXiv preprint arXiv:1612.01452}, 2016.

\bibitem{simonyan2014very}
K.~Simonyan and A.~Zisserman.
\newblock Very deep convolutional networks for large-scale image recognition.
\newblock {\em arXiv preprint arXiv:1409.1556}, 2014.

\bibitem{wang2016accelerating}
P.~Wang and J.~Cheng.
\newblock Accelerating convolutional neural networks for mobile applications.
\newblock In {\em Proceedings of the 2016 ACM on Multimedia Conference}, pages
  541--545. ACM, 2016.

\bibitem{wu2016quantized}
J.~Wu, C.~Leng, Y.~Wang, Q.~Hu, and J.~Cheng.
\newblock Quantized convolutional neural networks for mobile devices.
\newblock {\em IEEE Conference on Computer Vision and Pattern Recognition
  (CVPR)}, 2016.

\bibitem{zhang2015accelerating}
X.~Zhang, J.~Zou, K.~He, and J.~Sun.
\newblock Accelerating very deep convolutional networks for classification and
  detection.
\newblock {\em IEEE Transactions on Pattern Analysis and Machine Intelligence
  (TPAMI)}, 2015.

\end{thebibliography}
}

\end{document}